# Tracking Hand Hygiene Gestures with Leap Motion Controller

Rashmi Bakshi, Jane Courtney, Damon Berry, Graham Gavin

*Abstract*—The process of hand washing, according to the WHO, is divided into stages with clearly defined two handed dynamic gestures. In this paper, videos of hand washing experts are segmented and analyzed with the goal of extracting their corresponding features. These features can be further processed in software to classify particular hand movements, determine whether the stages have been successfully completed by the user and also assess the quality of washing. Having identified the important features, a 3D gesture tracker, the Leap Motion Controller (LEAP), was used to track and detect the hand features associated with these stages. With the help of sequential programming and threshold values, the hand features were combined together to detect the initiation and completion of a sample WHO Stage 2 (Rub hands Palm to Palm). The LEAP provides accurate raw positional data for tracking single hand gestures and two hands in separation but suffers from occlusion when hands are in contact. Other than hand hygiene the approaches shown here can be applied in other biomedical applications requiring close hand gesture analysis.

## I. INTRODUCTION

Hospital Acquired Infections (HAIs) have a significant impact on quality of life and result in an increase in health care expenditure. According to the European Centre for Disease Prevention and Control (ECDC), 2.5 million cases of HAIs occur in European Union and European Economic Area (EU/EAA) each year, corresponding to 2.5 million DALYs (Disability Adjusted Life Year) which is a measure of the number of years lost due to ill health, disability or an early death. [1] MRSA-Methicillin Resistant Staphylococcus Aureus is a common bacteria associated with the spread of HAIs. [2]

One method to prevent the cross transmission of these microorganisms is the implementation of well structured hand hygiene practices. The World Health Organization (WHO) has provided guidelines about hand washing procedures for health care workers. [3] Best hand hygiene practices have been proven to reduce the rate of MRSA infections in a health care setting. [4]

One challenge in dynamic healthcare environments is to ensure compliance with these hand hygiene guidelines and to evaluate the quality of hand washing. This is often done through auditing involving human observation. The hand washing process, however, is well structured and has particular dynamic hand gestures associated with each hand washing stage.

The assessment of the process may therefore be suited to automation. Existing technology includes the use of electronic counters and RFID badges to measure the soap usage and location based reminder systems to alert the workers about washing hands. [17, 18, 19] These systems have shown to improve the frequency of hand washing but they do not assess if the process of handwashing is compliant with the guidelines.

One potential approach is to use 3D gesture trackers and/ or a camera system to track fine hand movements and identify user gestures, provide feedback to the user or a central management system, with the overall goal being an automated tool that can ensure compliance with the hand washing guidelines. In advance of developing these systems, however, preliminary analysis on the hand washing process and a structured methodology was required.

The aim of this paper is to analyze the WHO hand washing stages, as performed by healthcare professionals, and extract the unique hand features related with each stage, in preparation for further processing.

Using this structured approach, preliminary gesture tracking experiments were performed with a 3D gesture tracking device- a Leap Motion Controller (LEAP) in a configuration as shown in Figure 1.These results are presented with a preliminary assessment of the system's suitability in automatically detecting the completion of hand washing stages.

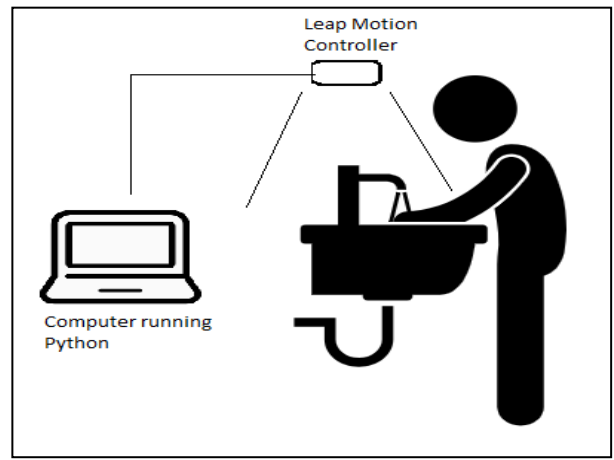

Figure 1: Diagram of Gesture Tracking System in Handwashing

## II. BACKGROUND

### A. WHO Hand Hygiene Stages

The World Health Organisation has clear guidelines for washing hands segmented into 11 sequenced stages. Stages 2-7 directly involve hand washing, and are the focus of this paper, with the other stages related to turning on water, drying hands etc. [3]. The overall washing time and the time spent at each stage is also an important factor for the analysis.

The hand washing process is clearly a two-handed dynamic process and this presents unique challenges for tracking systems. Challenges include hand occlusion and 'confusion', which occurs due to hands overlapping, intertwining each other and where one hand blocks the other. In addition, nearly all stages involve challenging dynamic features that help characterise gestures, including linear and rotational movement.

*B. Hand Features and Gestures*

When attempting to track hand gestures it is important to determine the specific features associated with each gesture. Such a process is useful in both sequential programming using thresholds and loops but also in potential machine learning applications.

A *feature* is characteristic information of the local appearance of the object to be tracked [7], for this application - the hands. As shown in Figure 2, they can include fingertip position vectors, angles between fingers, overall position in co-ordinate system, velocities etc. Other researchers have further included hand curvature and palm orientation for static single and two hand gestures. Feature extraction is an important step in building a hand gesture recognition system.

*Gestures* are comprised of a number of features. Features are selected in such a way that they have the ability to separate and identify a single gesture-class from the group of other gestures. It is proposed that a sequence of tracked gestures will then be used to determine that each hand washing stage has been completed.

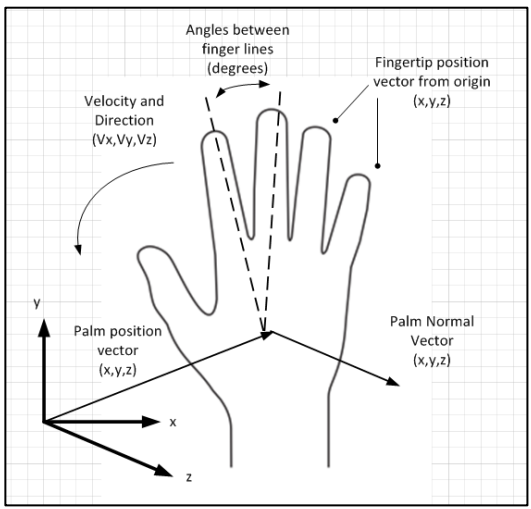

Figure 2: Example of hand features for gesture tracking

*C. Gesture Tracking*

The technology used for tracking gestures tends to be based on 1) image processing, 2) gesture tracking devices or 3) a combination of (1) and (2). 3D depth sensors have been widely used by many researchers in the field of gesture recognition to detect single handed gestures. [5, 6] They can track specific features on the hand such as each finger-tip positional vector or palm positional vector, for example. Most only provide raw positional data that needs further processing to extract useful features.

Previous work from researchers have focused on the specific challenges of tracking hands in motion.

Lu *et al.* (2016) explored palm vectors for the classification of dynamic single hand gestures [15]. Hamda *et al.* (2017) extracted hand contours and hand orientation from a static hand image acquired via a 3D sensor (Microsoft Kinect). The study was done for one hand gestures. [10] Using a LEAP, Lin Shao *(2015)* extracted the fingertip position and palm center for tracking various hand gestures [5]

Marin *et al.*(2015) also used LEAP and the Microsoft Kinect jointly to extract hand features such as finger-tip position, hand center and hand curvature to build up unique combinations for recognizing American sign language gestures. [6][9]

Liorca, Xia *et al.* built a hand gesture recognition system for the classification of six hand hygiene poses that involved two hands over the dataset of 72 RGB videos and the corresponding depth videos using soft kinetic camera. They extracted histogram of oriented gradient features on the hand regions and applied principal component analysis for the pose classification. [11] [12]

In biomedical and related disciplines previous researchers have used the LEAP for a range of applications. Szwedo *et al.* (2018) assessed the validity of the LEAP for performance analysis of upper limb visually guided movements. [20] Butt *et al.* (2017) investigated the LEAP for assessing motor dysfunction in patients with Parkinson's disease. [21]

## III. METHODOLOGY

*A. Observational feature extraction study*

In this study, following the approach in Bhuyan *et al.* the focus is on human recognizable features that can be applied to two hands [14]. Direct observation was used to analyze the hand washing video recordings of health care workers, in order to extract the unique features associated with each hand hygiene (HH) stage. Ten videos demonstrating WHO hand washing guidelines were sourced from health authority websites (link provided in *Results* section). It was also important to capture each individual expert's interpretation of each stage.

The corresponding human observed classification values for each feature in each hand washing stage was tabulated. A unique combination of these features could then be used to classify each individual stage. As an example, a comparison of features for Stage 2: Rub hands Palm to Palm and Stage 3: Right Palm over Left Dorsum and vice versa is shown in the Table 1. Once the unique features were identified, the next step was to determine if they could be detected with a 3D gesture tracking device- LEAP.

## B. Preliminary tests conducted with 3D sensor- Leap Motion Controller (LEAP)

The Leap Motion Controller (LEAP) is a low cost, low power USB peripheral device with two inbuilt monochromatic IR cameras and three infrared LEDs, designed for hand and arm tracking up to approximately 60cm (viewing range). [13]

The Leap Software Development Kit (SDK) allows access to the raw joint positional data, as well as some useful in-built hand features, such as Palm Normal Vector, Palm Position Vector and Hand Grab Strength (Hand Curvature Scalar). This positional data with additional programming implemented in Python, and some of the inbuilt functions, were used to extract hand features, determine gestures and assess the completion of hand hygiene (HH) stages. The sampling frequency for all the tests were within the range of 100-110 frames per second (FPS). Left-hand and Right-hand data samples were stored in a separate comma separated file (csv) file for all the preliminary experiments.

Preliminary experiments focused upon:
i. Determining the optimum location for LEAP to extract the relevant hand features during hand washing.
ii. Ability of LEAP to operate under flowing water from a faucet.
iii. Writing specific blocks of python code to extract the identified relevant single and two- hand features.
iv. Applying appropriate thresholds to these blocks to optimise the performance.

Following the preliminary experiments the overall methodology was applied to Stage 2: Rub hands Palm to Palm as a preliminary test to determine its tracking and assessment performance.

## C. Application to WHO Stage 2- Rub hands palm to palm

In order to track a complete stage, the blocks of code developed in the previous section can combined into a decision making tree and implemented in Python.

To demonstrate this, the **WHO Stage 2- Rub hands palm to palm**, see user instructional graphic in Table 1, was chosen as it included a number of challenging gestures in a particular sequence. This HH stage is the first time that the two hands are in contact and was the determining factor for the completion of Stage 2.

## D. Preparation for machine learning

The feature extraction method outlined above can also be applied to building a training set for gesture recognition and classification in machine learning. Previous researchers have used supervised learning algorithms to detect single hand gestures. Support Vector Machine (SVM) classifier has been widely applied for detection of American Sign Language and counting gestures [6, 15].

## IV. RESULTS

### A. List of hand features for identification of Hand Hygiene stages

All ten expert hand washing videos were assessed for human observable features. Despite the WHO instructional graphic for each stage the process is open to individual interpretation and this was noted also.

| WHO Stage 2,3 User Instruction Graphic | Features | WHO Stage 2 | WHO Stage 3 | Can classify stage 2 and 3? |
|---|---|---|---|---|
| 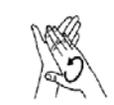 Stage 2 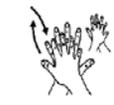 Stage 3 | Palm Orientation | Palms facing each other | One Palm over another palm | **Yes** |
| | Palm Shape | Flat | Flat | No |
| | Finger Spread | Closed | Open | **Yes** |
| | Hand Trajectory | Linear or Circular | Linear | No |
| | Frequency of movement | 0.8-3.6¬Hz | 1-3 Hz | No |
| | Time Duration | 2-7 sec | 1– 10 sec | No |

Table 1: **Comparison of** Unique hand features of WHO Stage 2 and 3- extracted through expert video analysis

Table 1 shows the feature comparison for Stages 2 and 3, although this process was completed for all stages. A database record of the human observations is maintained for all the other stages and can be accessed online alongside links to the associated public domain hand washing videos, see link:

| Open access link to database of hand washing stages |
|---|
| https://drive.google.com/drive/folders/11LswympjpDq3kA7HiYTDRlhO-I9TSFGu?usp=sharing |

After observation studies of all videos it was determined that the following features, in different combinations, could uniquely describe every HH stage: 1) Palm orientation, 2) Palm shape, 3) Finger spread, 4) Hand trajectory and 5) Rate of movement. Stage duration is not a feature but was included as an important identifier of successful stage completion.

In addition, distance and location features will inevitably be utilized to determine the relative position between the two handed gestures in HH stages. Beyond the individual's interpretation of stages, other actions were noted in the video footage, such as nail cleansing and washing wrists. These were not included in this study.

## B. Preliminary extraction of features using LEAP and Python

Prior to exploring feature extraction, it was important to determine the optimum positioning of the leap device and its performance under water in order to progress with tracking HH stages.

### LEAP position relative to hands

Ideally, LEAP works best when placed flat on a flat surface below the hands. All the preliminary tests were done keeping the sensor in that position. However, for the purpose of tracking hands in a hand washing process, it was important to determine the optimum location of LEAP for two-handed gestures.

| Position | Pros | Cons |
|---|---|---|
| Horizontally placed on the surface | Occlusion -Low Tracking - Excellent | Difficult to implement with a faucet |
| Facing on the side | Tracking – Good Easy to position around faucet | Occlusion – High (could be solved by multiple trackers) |
| Front facing | Occlusion -Low | Tracking – Hands abruptly lost. |

Table 2: LEAP positioning suitable for tracking HH gestures

### LEAP tracking under water

The LEAP sensor has the ability to maintain hand tracking in water, as shown in wire frame in Figure 3.

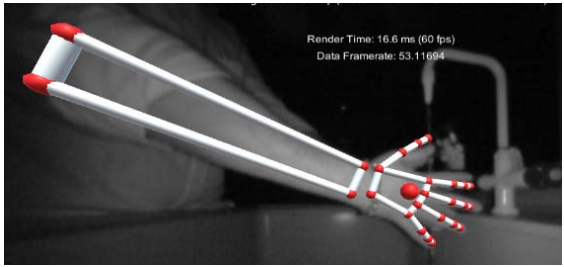

Figure 3: Wire frame of Hand Tracking with LMC under Water

### Tracking Hand Features

Each hand feature was assessed for methods to track and identify using the LEAP:

**Palm Orientation:** In all the HH stages, palms of two hands are 1) facing each other or 2) one palm over another palm (front to back). To determine if palms are either facing each other, the Palm Normal Vector (a unit vector that is orthogonal to the palm pointing outwards) was utilised. If palms are facing each other then:

Case 1) Same magnitude and opposite direction for Palm Normal Unit vectors.

$$\vec{A}=\vec{A}_x+\vec{A}_y+\vec{A}_z$$
$$\vec{B}=\vec{B}_x+\vec{B}_y+\vec{B}_z$$
$$\text{resultant vector}, \vec{C}=\vec{A}+\vec{B}$$

ABS RH (A) + LH (B) < 0.4 for palms facing each other else palms facing in one direction.

**Palm Shape:** The vendor-supplied Python function, Hand Grab Strength, was utilized to find out if each hand was flat or curved with 0 as completely flat and 1 as completely curved [16]. After running the experiments, it was found that for the purpose of hand washing application, hands were considered flat in the range 0-0.3.

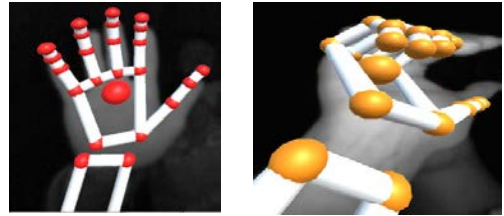

Figure 4: Flat and Curved palms

**Fingers Spread:** This feature corresponds to the finger position and orientation in a hand gesture. It can be used to compute the angles and distance between the fngers and the centre of the palm. In a wide open hand, the minimum distance between two finger tips was 17mm after conducting repetitive experiments. Finger-tip distance between two fingers can be further integrated with the palm shape to determine if the hand is fully open or closed.

**Hand Trajectory:** refers to the type of hand movement in a particular HH stage. From video analysis, it was noted that the hand trajectory for all stages is either linear or circular.

While all the important features could be tracked and processed, preliminary results highlighted the challenge of tracking two hands in contact. While the LEAP data for two hands in separation is excellent the LEAP suffered from occlusion when hands were in contact. This highlighted the importance of positioning the device to minimize occlusion. For tracking stages in the following section the LEAP was positioned horizontally flat to maximize the capture of 1 hand as in some cases this was sufficient to classify the HH stage. Other proposed solutions are:

- Using multiple LEAPs (only recently supported)/ gesture trackers to capture the hands from multiple angles.
- Use a combination of cameras and trackers.

## C. Automated detection of WHO Stage 2- Rub hands palm to palm

LEAP data for tracking hand features such as palm shape, palm orientation and fingers spread were integrated together along with number of hands present to detect the execution and completion of WHO Stage 2- Rub hands palm to palm. Figure 5 represents the flow of the Python code that detects the HH stage. The system tracks the presence of 2 hands, if they are facing each other. If not, it alerts the user. Then, it calculates the distance between two hands and checks if it decreases gradually as hands come in contact. In contact, number of hands is reduced to one as other hand is lost due to occlusion, it checks the palm velocity for a rotational movement and then concludes the execution of HH Stage2.

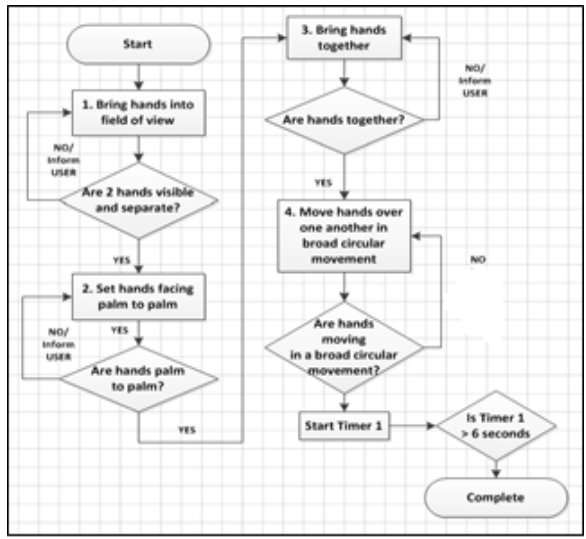

Figure 5: Flow chart of programming sequence for WHO Stage 2

*D. Proposed Feature Set for a machine learning model*

In preparation for future machine learning approaches hand features (and relevant two hand features) datasets are being prepared. Table 3 shows the overall dataset configuration.

| Sample No. | Hand Curvature LH, RH | Fingertip Distance LH,RH | Features …X n | Gesture Class |
|---|---|---|---|---|
| 1 | 0.6 | 25.66 | | Hands Palm to palm |
| 2 | … | … | … | Palm to Palm with fingers interlocked |
| Samples …. n | … | … | … | Label |

Table 3: Two hand Feature set for Hand Hygiene stages classification

## V. CONCLUSION

This paper has followed a structured methodology and presented the hand features associated with the WHO hand washing stages as extracted from healthcare workers. These features can be used by any tracking approach and also in preparation for machine learning methods.

In addition the suitability of a the 3D gesture tracker, LEAP Motion Controller, tracking hand gestures involved in hand hygiene stages was determined. The LEAP provides accurate positional data for hand and arm movements while the hands are in separation but suffers from occlusion when hands are in contact.

Even so, with appropriate positioning of the LEAP, it was possible to track a complex sample WHO stage where hands are in contact and moving by focusing on the actions of one hand.